\pdfoutput=1
\documentclass[letterpaper]{article} 
\usepackage[submission]{aaai24}  
\usepackage{times}  
\usepackage{helvet}  
\usepackage{courier}  
\usepackage[hyphens]{url}  
\usepackage{graphicx} 
\usepackage{natbib}  
\usepackage{caption} 
\usepackage{algorithm}
\usepackage{algorithmic}
\usepackage{amsmath}
\usepackage{amsfonts}
\usepackage{subcaption}
\usepackage{booktabs}
\usepackage{multirow} 
\usepackage{multicol}
\usepackage{amsthm}
 
\usepackage{newfloat}
\usepackage{listings}
\DeclareCaptionStyle{ruled}{labelfont=normalfont,labelsep=colon,strut=off} 
\floatstyle{ruled}
\newfloat{listing}{tb}{lst}{}
\floatname{listing}{Listing}

\title{Not All Data Matters: An End-to-End Adaptive Dataset Pruning Framework for Enhancing Model Performance and Efficiency}
\author{Suorong~Yang\textsuperscript{\rm 1, \rm 2}, 
	Hongchao~Yang\textsuperscript{\rm 1, \rm 2}, 
	Suhan~Guo\textsuperscript{\rm 1, \rm 3}, 
	Furao~Shen\textsuperscript{\rm 1, \rm 3 \thanks{Corresponding author.}},
        and Jian~Zhao\textsuperscript{\rm 4},  }
\affiliations{
    \textsuperscript{\rm 1}National Key Laboratory for Novel Software Technology, Nanjing University, China\\
    \textsuperscript{\rm 2}Department of Computer Science and Technology, Nanjing University, China\\
    \textsuperscript{\rm 3}School of Artificial Intelligence, Nanjing University, China\\
    \textsuperscript{\rm 4}School of Electronic Science and Engineering, Nanjing University, China\\
    sryang@smail.nju.edu.cn, yanghc@smail.nju.edu.cn, shguo@smail.nju.edu.cn, frshen@nju.edu.cn, jianzhao@nju.edu.cn
}
\usepackage{bibentry}

\begin{document}
\newtheorem{proposition}{Proposition}
\maketitle

\begin{abstract}
While deep neural networks have demonstrated remarkable performance across various tasks, they typically require massive training data.
Due to the presence of redundancies and biases in real-world datasets, not all data in the training dataset contributes to the model performance.
To address this issue, dataset pruning techniques have been introduced to enhance model performance and efficiency by eliminating redundant training samples and reducing computational and memory overhead.
However, previous works most rely on manually crafted scalar scores, limiting their practical performance and scalability across diverse deep networks and datasets.
In this paper, we propose \textbf{AdaPruner}, an end-to-end \textbf{A}daptive \textbf{DA}taset \textbf{PRUN}ing fram\textbf{E}wo\textbf{R}k.
AdaPruner can perform effective dataset pruning without the need for explicitly defined metrics. 
Our framework jointly prunes training data and fine-tunes models with task-specific optimization objectives.
AdaPruner leverages (1) An adaptive dataset pruning (ADP) module, which iteratively prunes redundant samples to an expected pruning ratio; and (2) A pruning performance controller (PPC) module, which optimizes the model performance for accurate pruning.
Therefore, AdaPruner exhibits high scalability and compatibility across various datasets and deep networks, yielding improved dataset distribution and enhanced model performance.
AdaPruner can still significantly enhance model performance even after pruning up to 10-30\% of the training data.
Notably, these improvements are accompanied by substantial savings in memory and computation costs.
Qualitative and quantitative experiments suggest that AdaPruner outperforms other state-of-the-art dataset pruning methods by a large margin.
\end{abstract}
\section{Introduction}
Deep learning has made remarkable progress in image classification and various computer vision tasks.
Despite this tremendous progression, training deep networks to achieve satisfactory performance typically necessitates a considerable volume of data due to the high Vapnik–Chervonenkis (VC) dimension inherent to deep networks. 
Unfortunately, data collected in the real world often suffer from redundancies and biases.
As empirically demonstrated in~\cite{data_diet}, the redundancy in datasets increases the risk of overfitting.
Meanwhile, it has been observed that beyond a certain threshold, augmenting the volume of data no longer yields improved model performance or even hurts it~\cite{lou2020deep,hein2019relu}.
Consequently, excessive increases in data volume may impede model generalization~\cite{double_decent}, meaning that not all training data contribute beneficially to the model performance. 
Moreover, utilizing large datasets also incurs substantial memory and computation overhead.

To address the challenge and harness the dataset more effectively, recent efforts have focused on dataset pruning techniques.
Dataset pruning removes redundant or irrelevant data from a training dataset to prepare a pruned dataset for subsequent tasks.
The pruned datasets enable deep models to focus on meaningful patterns, reduce potential biases,  and ultimately enhance model performance while reducing computational and memory costs. 
Previous works have attempted to prune redundant training data using some hand-crafted scalar scores~\cite{herding,core-set3}, such as Forgetting~\cite{forgetting}, GraNd~\cite{data_diet}, CG-Score~\cite{cgscore}, etc.
However, their scalability across different deep networks and datasets is limited, primarily due to their reliance on predefined selection standards.
For instance, Forgetting and CG-score define importance scores based on supervised image classification tasks, rendering them inapplicable to object detection datasets.
Additionally, the group effect of the combination of low-score and high-score samples, possibly contributing to model performance, is neglected by these score-based approaches.
In fact, we have proved theoretically that the dataset pruning problem is NP-hard (Proposition~\ref{prop:1}), indicating that \textbf{it is challenging to find a perfect scoring standard that can work well across various models and datasets}.
Hence, an intriguing and vital question arises: \textit{Is it possible to teach models to determine which samples should be pruned?}

To address the challenges above, we propose \textbf{AdaPruner}, an end-to-end \textbf{A}daptive \textbf{DA}taset \textbf{PRUN}ing fram\textbf{E}wo\textbf{R}k, a simple yet powerful mechanism to adaptively identify and remove unimportant samples across deep networks and datasets.
Unlike most previous approaches, the proposed framework avoids explicitly defining the importance score for each sample.
AdaPruner integrates the dataset pruning optimization objective into the task-specific optimization process, enabling adaptive pruning.
During each epoch, it performs adaptive dataset pruning (ADP) by penalizing the redundant samples, with a pruning performance controller (PPC) module maintaining model performance for sample selection.
This highlights the robust scalability of AdaPruner as it can be applied to any loss-based deep networks, including supervised image classification and object detection datasets, etc.
Notably, our method effectively mitigates the group effect by enabling model-driven sample pruning.
Upon obtaining pruned datasets, they will be employed in downstream tasks without any training or inference from AdaPruner.
Remarkably, even with a 10-30\% reduction in data, models trained on the pruned datasets consistently outperform those using the original complete datasets.
Additionally, the pruned datasets also contribute to significant reductions in memory and computational overheads for downstream tasks.
To verify the effectiveness of AdaPruner, we conduct experiments on CIFAR-10/100~\cite{cifar-10,cifar100} and Tiny-ImageNet~\cite{tiny} for image classification, PASCAL VOC2007 and VOC2012~\cite{voc} for object detection, and MNIST~\cite{mnist} for visualization.
Experimental results demonstrate the strong scalability and state-of-the-art (SOTA) pruning performance achieved by AdaPruner, outperforming both score-based and non-score-based methods.
In summary, we highlight our contributions as follows:
\begin{itemize}
	\item We propose AdaPruner, an end-to-end adaptive dataset pruning framework that enables automatic pruning to any expected ratio without dependence on any hand-crafted scalar scores.
        \item Theoretically, this study establishes the complexity of the general dataset pruning problem by demonstrating that it is NP-hard.
	\item Our framework does not impose any assumptions on datasets or models, rendering its strong scalability to any datasets and loss-based deep networks.
	\item Through extensive experiments on various benchmark datasets, we validate AdaPruner's effectiveness in improving the distribution of training datasets and showcase its superiority in comparison to other SOTA dataset pruning methods.
\end{itemize}
\section{Related Work}
The primary goal of dataset pruning is to reduce the size of training data while minimizing the impact on model performance for subsequent tasks~\cite{dataset_pruning}.
The concept of dataset pruning originates from coreset selection, which involves selecting a small subset of highly informative training samples from a large dataset~\cite{deepcore,yoon2020data}.
Coreset selection typically involves sorting and selecting a portion of data based on predefined scalar scores for individual samples~\cite{forgetting, scail, herding, class_center}.
Models trained on the coreset will exhibit comparable performance to those trained on the complete training set.
For instance, Herding~\cite{herding}, a geometry-based method, selects data based on the distance between samples and the corresponding class centers in the feature space.
Forgetting~\cite{forgetting}, a loss-based method, monitors transitions from correct to incorrect classification, which is defined as the forgetting event.
 The frequency of forgetting events of each sample reveals the intrinsic properties of the training data, and the accuracy is not affected by training samples that are rarely forgotten.
Therefore, those samples can be removed.
GraNd and EL2N~\cite{data_diet} leverage loss gradient norms and $l2$-distance of the normalized error between the predicted probabilities and one-hot labels of the individual training samples to identify a subset of training data.
Specifically, the GraNd score of a training sample $(x,y)$ at time $t$ is defined as $\chi_t(x, y)=\mathbb{E}_{\mathbf{w}_t}\left\|g_t(x, y)\right\|_2$, where $g_t(x,y)$ is the gradient norm, which is defined as $g_{t}(x, y)=\nabla_{\mathbf{w}_{t}} \ell\left(p\left(\mathbf{w}_{t}, x\right), y\right)$.
The EL2N score of a training sample $(x,y)$ is defined as $\mathbb{E}\left\|p\left(\mathbf{w}_t, x\right)-y\right\|_2$, which is the $l2$-norm of the error vector.
Glister~\cite{glister} formulates the problem as a mixed discrete-continuous bi-level optimization, which selects a subset of the training data that maximizes the log-likelihood on a hold-out validation set.
Recently, ~\cite{cgscore} introduces a complexity-gap score to assess the influence of individual instances, which quantifies the irregularity of the instances and measures how much each data instance contributes to the overall improvement of the network training.
However, those methods are task-specific, and the effectiveness of those methods in deep learning is under debate~\cite{deepcore,dataset_pruning}.
~\cite{beyond} also observes high redundancy among training samples and proposes a scalable metric to quantify the significance of each sample.
Inspired by the influence function~\cite{influence-function} in statistical machine learning, ~\cite{dataset_pruning} proposed an optimization-based sample selection method by examining the influence of removing a specific set of training samples on models' parameters.
In addition, ~\cite{core-set,core-set2, glister,core-set4} define the problem of active learning as the coreset selection, where a model trained over the selected subset can generalize well on the remaining data.  
However, those techniques make conservative estimates, resulting in poor practical performance~\cite{data_diet}. 
Meanwhile, different from active learning's online re-weighting of samples, dataset pruning works offline, preparing pruned datasets for subsequent applications.
\begin{figure}[]
	\centering
	\includegraphics[width=0.5\textwidth]{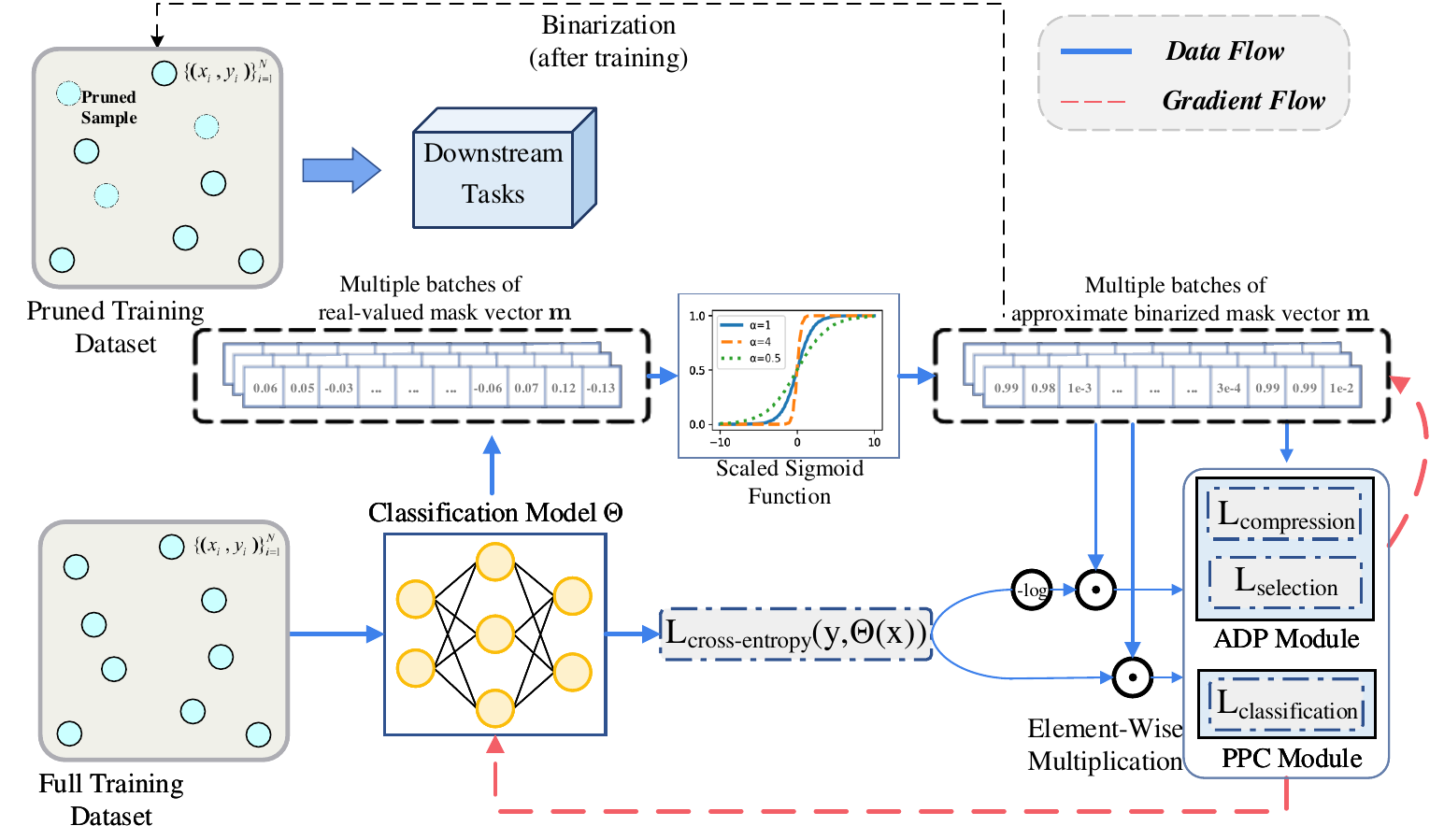}
	\caption{Overall framework of the proposed AdaPruner.}
	\label{framework}
\end{figure}
\section{AdaPruner}\label{sec:proposed-pruner}
\paragraph{Preliminary} The objective of dataset pruning is to prune dataset $\mathcal{D}$ into a pruned dataset $\hat{\mathcal{D}}^-$, containing no more than $k$ samples without sacrificing model performance.
Therefore, the overarching aspiration is for the test performance derived from the pruned datasets to closely approximate or even surpass the test performance achieved using the original full dataset, which can be formulated as follows:
\begin{equation}\label{eq:problem}
 \operatorname*{arg\,min}_{\hat{\mathcal{D}}^- \subseteq \mathcal{D},|\hat{\mathcal{D}}^-|\leq k } \mathbb{E}_{z \sim P(\mathcal{D})}\left[ L\left(z, \hat{\theta}_{\hat{\mathcal{D}}^-}\right)\right] -  \mathbb{E}_{z \sim P(\mathcal{D})}\left[ L(z, \hat{\theta})\right] 
\end{equation}
where $L(\cdot)$ denotes the loss function, $\mathcal{D}$ is a dataset, $P(\mathcal{D})$ represents the data distribution, $z$ is a test sample, and $\hat{\theta}$ and $\hat{\theta}_{\hat{\mathcal{D}}^-}$ are the empirical risk minimizers on the training set and pruned training set, respectively, $\textit{i.e.}$, $\hat{\theta}=\arg \min _{\theta \in \Theta} \frac{1}{n} \sum_{z_i \in \mathcal{D}} L\left(z_i, \theta\right)$ and $\hat{\theta}_{\hat{\mathcal{D}}^-}=\arg \min _{\theta \in \Theta} \frac{1}{n} \sum_{z_i \in \hat{\mathcal{D}}^-} L\left(z_i, \theta\right)$.
\begin{proposition}
\label{prop:1}
The dataset pruning problem in Eq.~\eqref{eq:problem} is NP-hard.
\end{proposition}
The proof is provided in the Appendix.
Proposition~\ref{prop:1} shows that finding an optimal solution to the dataset pruning problem is difficult.
Further compounding the challenge is the difficulty in predicting the generalization performance of the pruned dataset $\hat{\mathcal{D}}^-$ due to the unavailability of the test set during the training phase.
However, given that the training set and test set share the same data distribution $ P(\mathcal{D})$, the empirical risk serves as a viable surrogate for approximately estimating the performance of $\hat{\mathcal{D}}^-$.

As depicted in Figure~\ref{framework}, taking classification models as an example, our dataset pruning framework is modeled as an optimization process composed of two modules: Adaptive Dataset Pruning (ADP) and Pruning Performance Controller (PPC).
The objective of ADP is to iteratively minimize the discrepancy between practical and expected pruning rates by removing redundant samples.
 PPC operates collaboratively with ADP, iteratively minimizing the empirical risk associated with the set of samples selected by ADP, thereby facilitating precise sample selection.
The sample selection is indicated by the mask vector $\boldsymbol{m}$ within the ADP module.
Notably, because irregular batch sizes could affect resource utilization and optimization stability~\cite{batchsizes}, the selection of training samples in Eq.~\eqref{eq:PPC} of PPC and Eq.~\eqref{eq:ADP} of ADP is simulated by suppressing the gradient backpropagation of the pruned samples.
Upon completion of training, we obtain the pruned datasets for downstream tasks. 
The training of our framework utilizes the whole dataset.
Once the pruned dataset is obtained, the downstream tasks can directly employ the pruned datasets without any training or inference from AdaPruner, highlighting its efficiency.

An appealing aspect of our approach is that the pruning in each iteration is adaptively determined by model $\Theta$.
After each pruning step, the ADP module checks the gap between the practical and expected pruning ratios until it is below a predefined tolerance level $\epsilon$.
In this way, our framework achieves adaptive dataset pruning.
 \subsection{Warm-Up Mechanism and Pruning Performance Controller (PPC)}\label{warm-up-regime}
In the initial stage, all samples are selected for training.
However, commencing the pruning process from scratch may lead to inaccurate sample selection, especially during the early stages when models have not yet been adequately trained.
To address the issue, we introduce a warm-up mechanism whereby the models undergo initial training to attain satisfactory capabilities before the actual pruning, which aids in achieving precise dataset pruning. 

As the dataset pruning progresses, the amount of data used for model backpropagation optimization gradually decreases, potentially affecting the model's performance in sample selection.
To address the issue, the PPC module is utilized to assure the models' generalization performance for pruning samples.
The PPC is seamlessly integrated with ADP; these two components are jointly conducted and mutually reinforce each other in an adversarial manner, preventing imprecise data selection attributed to the degradation of model performance.
It is noteworthy that this conjoint training mechanism incurs little time overhead relative to the training of a classification model on the full dataset.
The objective function of PPC at time $t$ ($t$=1,2,...,$T$) can be formulated as the following optimization problem:
\begin{equation}\label{eq:PPC}
\hat{\theta}_t=   \arg \min_{\theta} \sum_{z\in \hat{\boldsymbol{m}}_{t-1}\odot \mathcal{D}}L(z, \theta_{t-1})
\end{equation}
Here $\hat{\boldsymbol{m}}_{t-1}\odot \mathcal{D}$ approaximately represents the selected samples at time $t-1$.
The PPC module focuses on the current model performance and mainly utilizes the selected samples at time $t-1$ to attain local empirical risk minimization, guided by our loss function $L$.

\subsection{Adaptive Dataset Pruning (ADP)}~\label{sec:ADP}
ADP aims to iteratively optimize parameter $\boldsymbol{m}$, which will be used to explicitly select samples in $\mathcal{D}$.
Elements of $\boldsymbol{m}$ possessing binary values, specifically 0 or 1, determine the selection status of a sample.
However, the binarized vectors are difficult to optimize in neural networks due to the lack of gradient calculation.
 To overcome this, we employ scaled sigmoid functions~\cite{autopruner} to approximate a binary mask denoted as $y = \mathrm{sigmoid}(\alpha \boldsymbol{m})$,  where $\alpha$ is a hyper-parameter.
 An increase in $\alpha$ engenders elements in $\boldsymbol{m}$ to converge more intimately towards either 0 or 1.
 When $\alpha$ becomes sufficiently large, the mask elements are approximately 0 or 1.
 To obtain more accurate pruning results during training, the value of $\alpha$ is incrementally elevated, thereby compelling the scaled sigmoid function to yield a binary mask apt for pruning.
 Post-training, $\mathbf{m}$ will be strictly binarized to indicate the sample selection explicitly.
 While our approach compels the model to output a mask with the desired compression ratio, the eventual compression ratio is determined dynamically. 
 As such, we require that the output mask falls within a tolerance range of $2\%$ from the expected compression ratio (i.e., $\epsilon=2\%$). 
  For instance, given a compression ratio of $80\%$, the resultant compression ratio would span between $78\%$ and $82\%$.
  Conclusively, the objective function of ADP can be formulated as the following optimization problem:
  \begin{equation}\label{eq:ADP}
   \hat{\boldsymbol{m}}_t=   \arg \min_{\boldsymbol{m}} \sum_{z\in \hat{\boldsymbol{m}}_{t-1}\odot \mathcal{D}}L(z, \hat{\theta}_{t-1})
  \end{equation}
  Here ADP adaptively prunes redundant samples through empirical risk minimization, guided by the loss function $L$.

  \subsection{Loss Function}\label{sec:loss-function}
 The proposed loss function is meticulously designed to facilitate adaptive and efficient dataset pruning within PPC and ADP.
Firstly, to accomplish the objective stated in Eq.~\eqref{eq:PPC}, the cross-entropy loss $L_{\text{CE}}$ is initially employed to compute the classification loss, denoted as $L_{\text{classification}}$, for the reserved samples:
 \begin{equation}
     L_{\text{classification}} = \frac{\mathrm{sigmoid}(\alpha\mathbf{m}) \cdot L_{\text{CE}}(y,\Theta(x))}{N} 
 \end{equation}
 Since $\mathrm{sigmoid}(\alpha\mathbf{m})$ is approximately binary, the gradient backpropagation of the pruned samples is suppressed, effectively marginalizing their influence.
 Consequently, models' updates primarily capitalize on the selected samples. 
 Notably, the $L_{\text{CE}}$ term can be easily modified to other \textbf{task-specific loss functions}, such as cross-entropy loss and regression loss, in the context of object detection tasks.
 \begin{figure*}[h]
 	\centering
        \begin{subfigure}[]{0.45\textwidth}
 		\centering
 		\includegraphics[width=8cm]{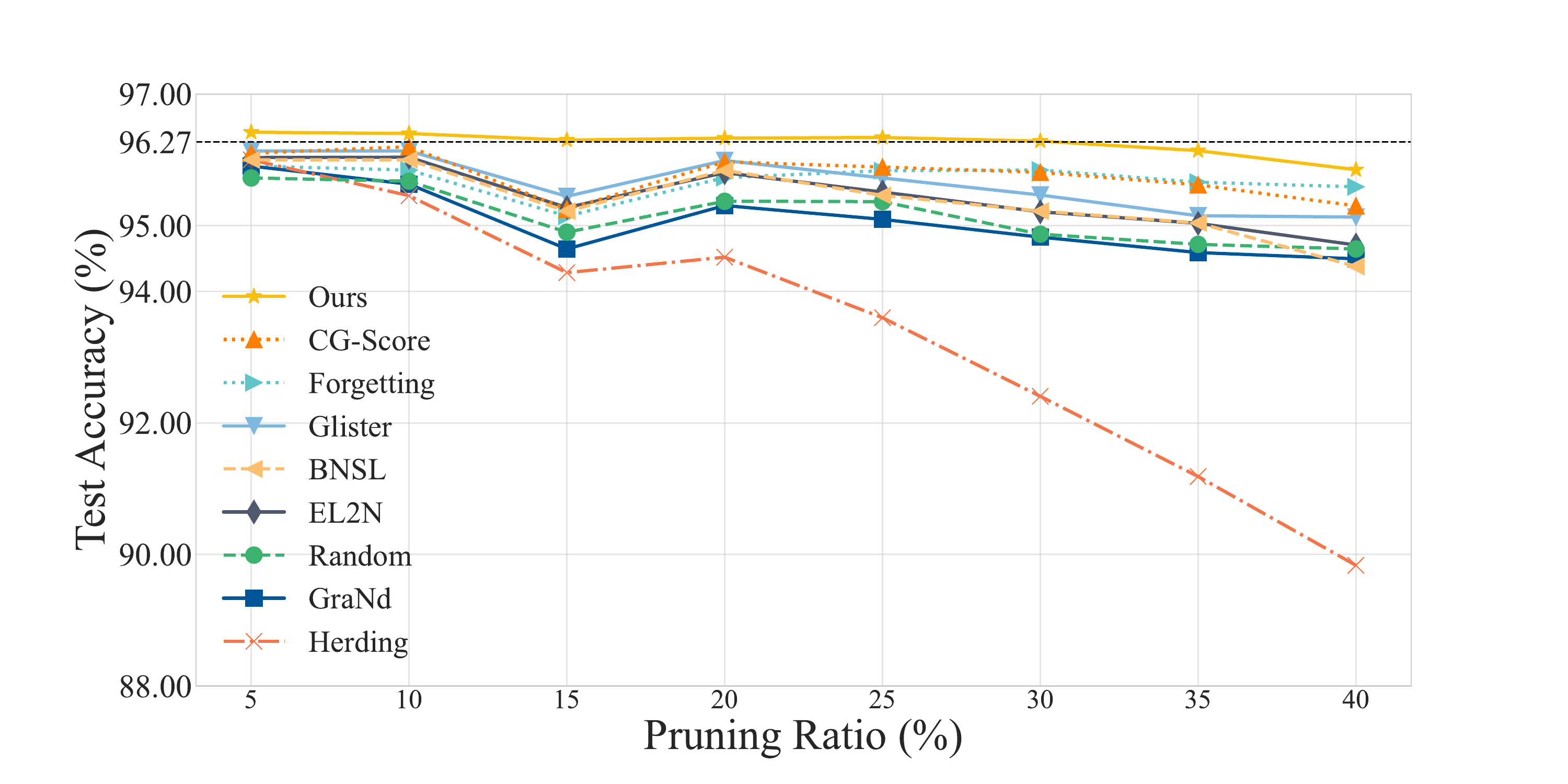}
 		\caption{ResNet-18 trained on CIFAR-10.}
 		\label{fig3-1}
 	\end{subfigure}%
 	\begin{subfigure}[]{0.45\textwidth}
 		\centering
 		\includegraphics[width=8cm]{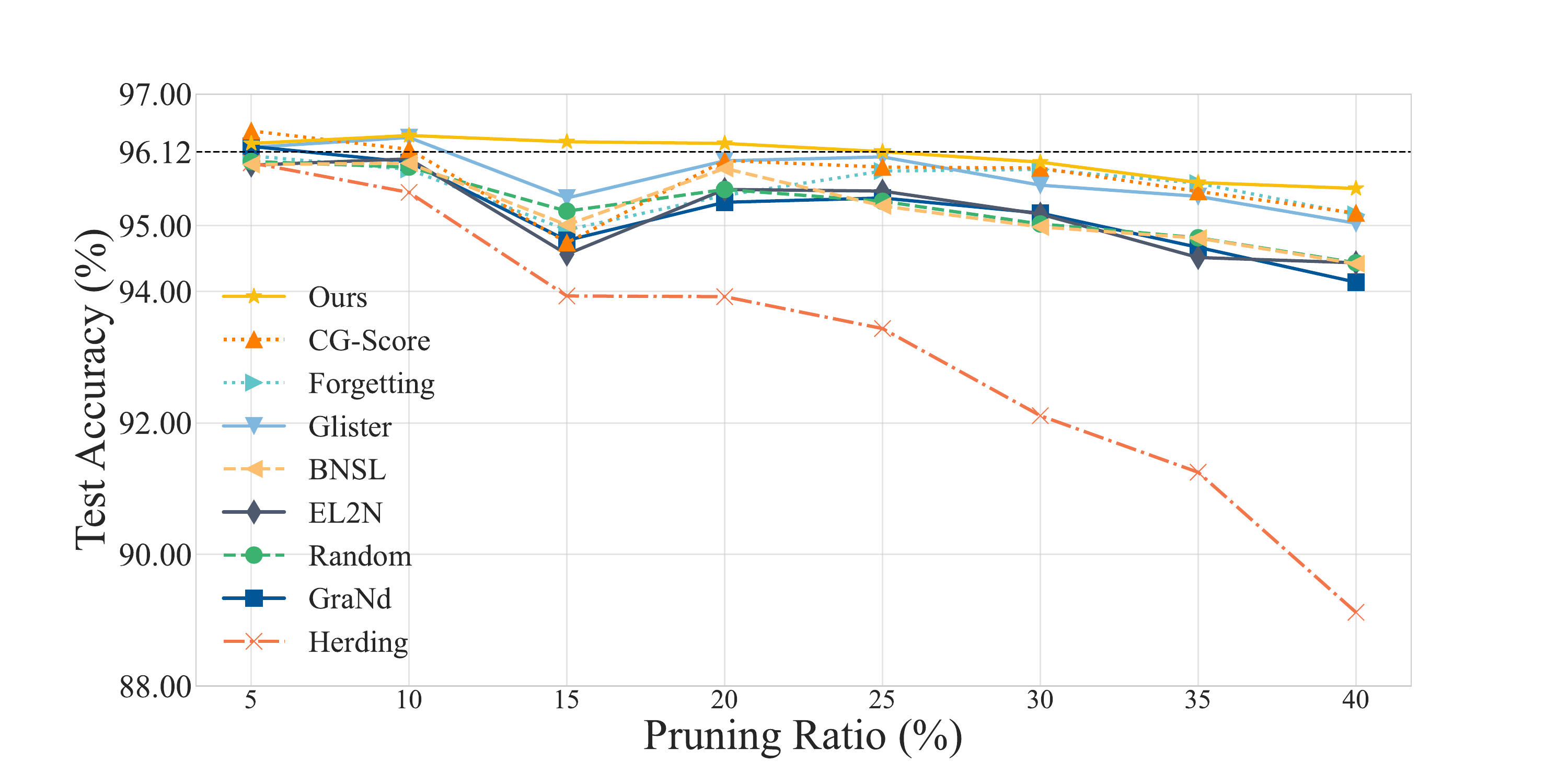}
 		\caption{ResNet-50 trained on CIFAR-10.}
 		\label{fig3-2}
 	\end{subfigure}
 	
 	\begin{subfigure}[]{0.45\textwidth}
 		\centering
 		\includegraphics[width=8cm]{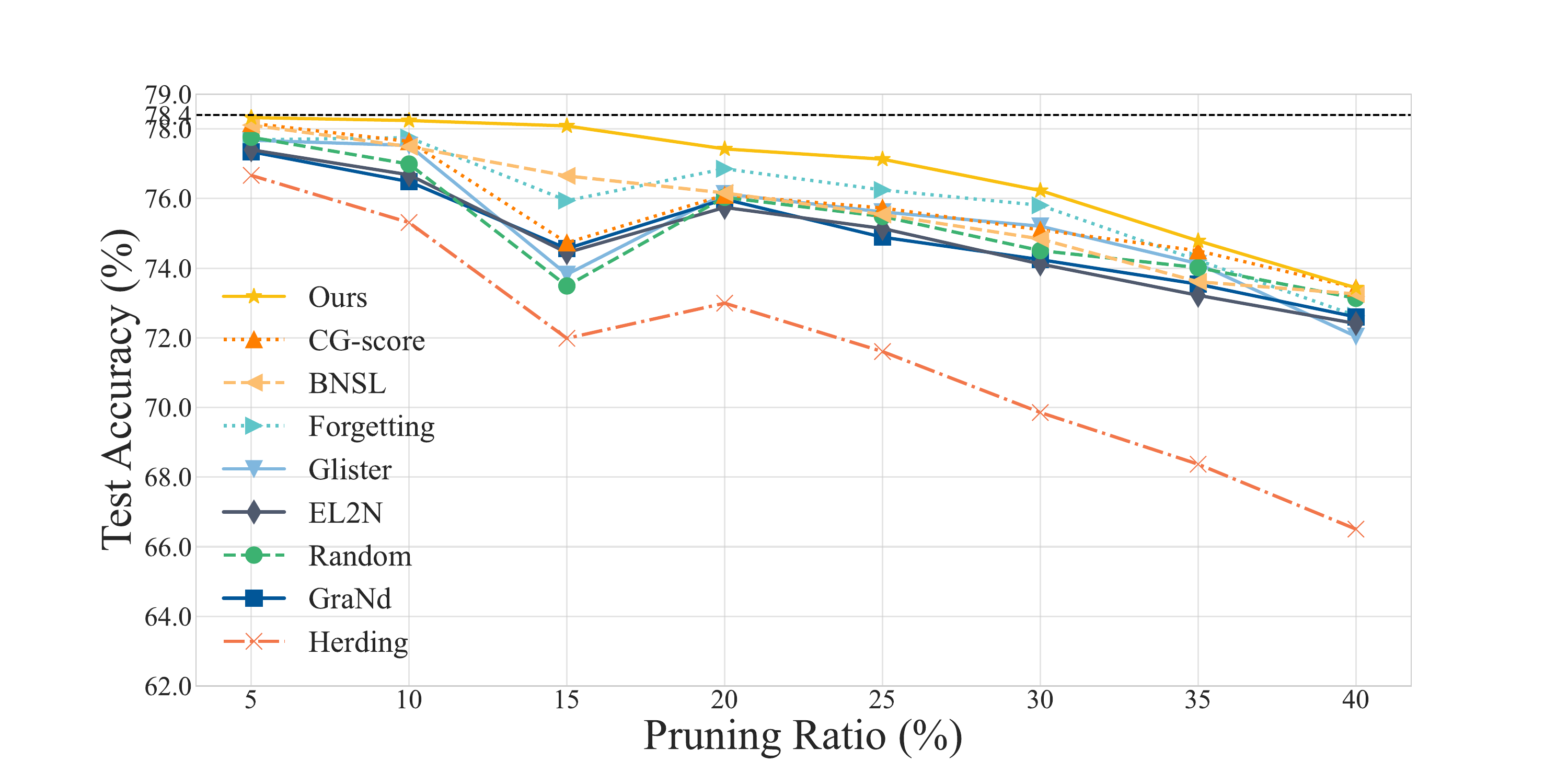}
 		\caption{ResNet-18 trained on CIFAR-100.}
 		\label{fig3-3}
 	\end{subfigure}%
 	\begin{subfigure}[]{0.45\textwidth}
 		\centering
 		\includegraphics[width=8cm]{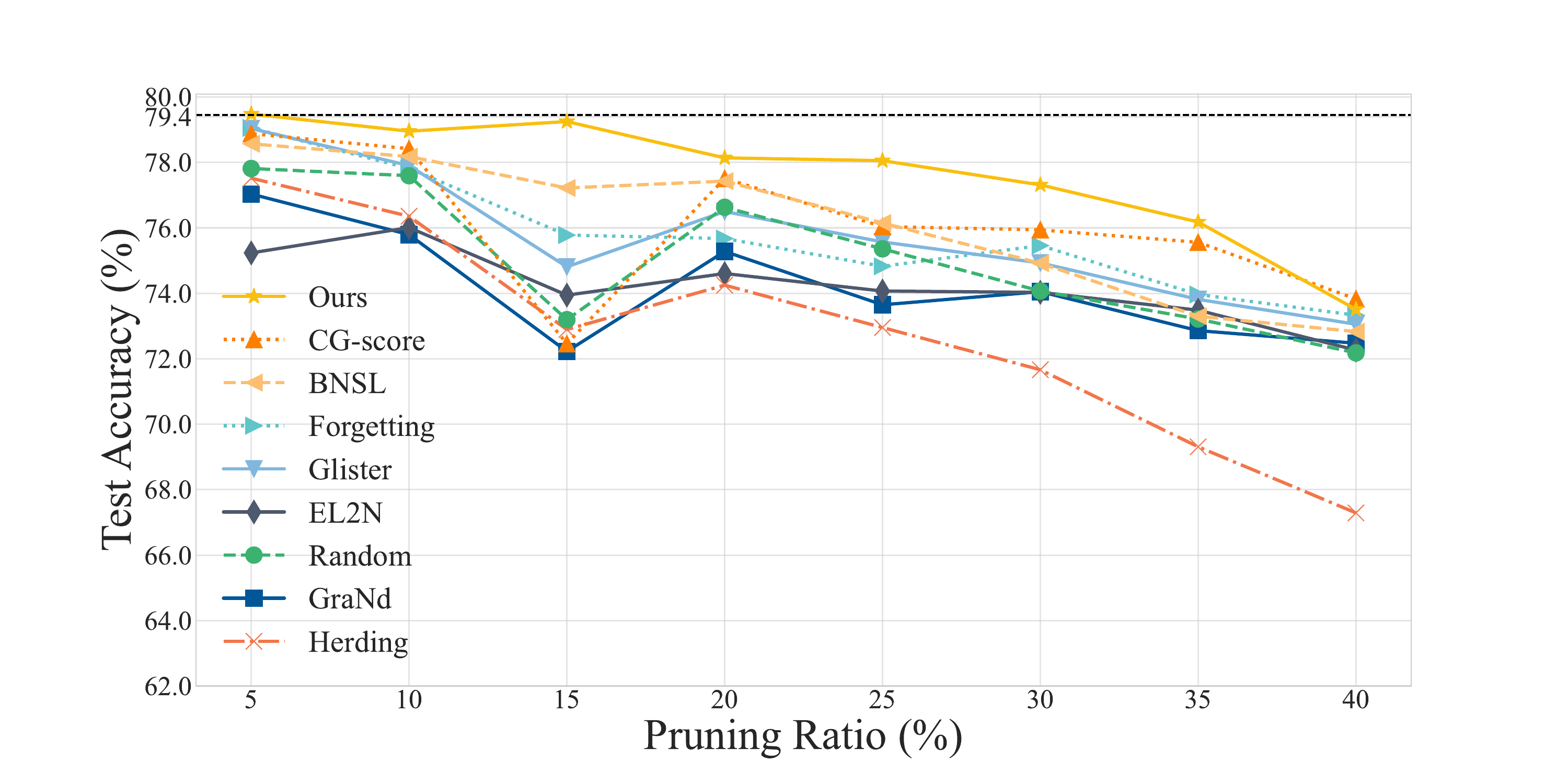}
 		\caption{ResNet-50 trained on CIFAR-100.}
 		\label{fig3-4}
 	\end{subfigure}
 	\caption{
 		ResNet-18 and ResNet-50 models are trained using pruned datasets by various pruning techniques on both CIFAR-10 (top) and CIFAR-100 (bottom).
        The black dashed line represents the test accuracy of training on the original full training set. The pruning ratio corresponds to the proportion of data removed.}
 \label{fig-accuracy}
\end{figure*}

Secondly, to enhance the performance of the ADP module during training, we introduce a data selection loss term.
Previous works on dataset pruning~\cite{dataset_pruning, forgetting} and curriculum learning~\cite{curriculum_learning} highlights that easy samples contribute less to the final test accuracy since they are less likely to be misclassified once correctly classified.
Hence, to encourage our method to prune simpler samples by updating $\mathbf{m}$, the data selection loss can be formulated as follows:
\begin{equation}
L_{\text{selection}} = \frac{\mathrm{sigmoid}(\alpha\mathbf{m}) \cdot [-log(L_{\text{CE}}(y,\Theta(x)))]}{N}
\end{equation}
Although $L_{\text{selection}}$ and $L_{\text{classification}}$ may exhibit similarities, $L_{\text{selection}}$ is primarily used to update the parameter $\mathbf{m}$ for sample selection, while $L_{\text{classification}}$ updates models via the gradient backpropagation of the chosen samples, thereby simulating sample selection.

To ensure that the models adaptively prune datasets to the expected compression ratio, a compression loss term is integrated. 
The scaled sigmoid function approximates the mask $\mathbf{m}$ to a binary vector.
Since the output values of $\mathbf{m}$ are approximately 0 and 1, the convex $l1$-relaxation of $\mathbf{m}$ can be used to represent the number of ones in  $\mathbf{m}$ as $\frac{||\textbf{$\mathbf{m}$}||_1}{N}$, where $N$ is the total number of the training samples.
Given the expected compression ratio $c_r$, the compression loss term can be defined as:
\begin{equation}
L_{\text{compression}} = \left[\frac{\left\|\mathrm{sigmoid}(\alpha\textbf{m})\right\|_1}{N}-c_r\right]^2
\end{equation}
Finally, our loss function is defined as follows:
\begin{equation}\label{loss-function}
L = L_{\text{classification}} + L_{\text{selection}} + \lambda L_{\text{compression}}
\end{equation}
where $\lambda$ is an adaptively adjusted coefficient that controls the relative importance of the actual pruning ratio.
It is updated based on the current actual compression ratio, i.e., $\lambda = s|r-c_r|$,
where $s$ is a scaling factor to adjust the numeric disparity among different loss items, $r$ is the current actual compression ratio during training, and $c_r$ is the expected compression ratio.
If the current compression ratio deviates far from the expected one, the model will prioritize the compression loss over the model fine-tuning; otherwise, more emphasis is placed on fine-tuning. 
Each component of the loss function is critical, and excluding any term would negatively impact the performance. 

\textbf{Complexity of AdaPruner} We present a theoretical analysis to show that our method does not involve notable computational overhead.
The computational complexities of $L_{\text{classification}}$ and $L_{\text{selection}}$ are $O(C \times N +N)$, where $N$ denotes the number of samples and $C$ is the number of categories.
The computational complexity of $L_{\text{compression}}$ is $O(N)$.
Consequently, the overall computational complexity is $O((C+1)\times N)$, nearly equivalent to the conventional cross-entropy loss.
The results of the practical training costs of our method are elaborated in the Appendix.

\section{Experiment}\label{Sec:experiment}
This section demonstrates AdaPruner's strong performance and analyzes its behavior through ablation studies.

\textbf{Datasets and Experimental Settings}
We evaluate the effectiveness of our method on six different benchmark datasets.
First, we present the experimental results of supervised image classification on widely used benchmarks, e.g.,  CIFAR-10/100~\cite{cifar-10} and Tiny-ImageNet~\cite{tiny}.
Second, we employ transfer learning to compare the performance of various dataset pruning methods. 
Third, we apply AdaPruner to the object detection tasks on PASCAL VOC2007 and PASCAL VOC2012~\cite{voc} datasets.
In addition, some analytical experiments are conducted to provide additional insights into our approach.
Specifically, to demonstrate that AdaPruner improves the training dataset's distribution and encourages models to learn better representation, we visualize the pruning effectiveness on MNIST~\cite{mnist}.
Ablation studies are also conducted to show how the components affect the pruning performance.

\textbf{Comparison with State-of-the-arts} 
We compare our method with 8 dataset pruning methods, including random pruning (Random), Forgetting~\cite{forgetting}, Glister~\cite{glister}, EL2N, GraNd~\cite{data_diet}, Herding~\cite{herding}, BNSL~\cite{beyond}, and CG-score-based method~\cite{cgscore}.

\textbf{Parameter Settings} 
The scaling parameter $s$ is set to 100.
Additionally, we adjust $\alpha$ used in ADP by specifying its initial and final values to prevent $\alpha$ from increasing too slowly or quickly. 
Specifically, $\alpha$ increases uniformly with training epochs. 
Our observations suggest that this strategy is governed by two factors. Firstly, the initial value of $\alpha$ controls the rate of binarization, provided that it is not too large. 
Larger initial values result in faster binarization but may degrade the performance of our method. Therefore, we typically set $\alpha$ to a value less than 1, such as 0.1. Secondly, if the mask is challenging to converge to binary values, increasing the maximal value of $\alpha$ appropriately can force the mask to converge towards binary values. 
On CIFAR10/100, we set $\alpha$ initially to 0.1 and increase it uniformly to 5 during training. On Tiny-ImageNet, $\alpha$ starts at 0.1, culminating at 100.

\textbf{Implementation Details}
We closely follow previous works~\cite{cutout, forgetting} with our setup.
Specifically, images are preprocessed by dividing each pixel value by 255 and normalized by the dataset statistics.
We train classification models using pruned datasets for 200 epochs with a batch size of 256 and a 0.1 learning rate with cosine annealing learning rate decay strategy, SGD optimizer with the momentum of 0.9, and weight decay of $5e^{-4}$. Data augmentation of random crop and random horizontal flip is added.
To decrease the risk of overfitting on CIFAR-10, Cutout~\cite{cutout} is applied across all methods.
\subsection{Supervised Image Classification}
\subsubsection{CIFAR-10 and CIFAR-100 Results}\label{sec:cifar}
To evaluate the performance of the pruned datasets, in Figure~\ref{fig-accuracy}, we conduct experiments on CIFAR-10/100 using pruning ratios ranging from 5\% to 40\%.  
It can be observed that our method consistently outperforms other methods on both CIFAR-10/100 using ResNet-18 and ResNet-50.
Notably, a substantial number of samples can be omitted from the training set while maintaining the models' generalization performance.
For CIFAR-10 with pruning ratios below 30\% and CIFAR-100 below 15\%, \textbf{the performance of our pruned datasets surpasses that of the entire dataset}, demonstrating the efficacy of AdaPruner in obtaining a smaller yet better training subset.  
 With higher pruning ratios, AdaPruner incurs \textbf{the least loss in accuracy} compared to other SOTA methods, showing enhanced model performance.
 
 Since all the pruned datasets are obtained using ResNet-18, Figure~\ref{fig3-2} and Figure~\ref{fig3-4} highlight the model-agnostic nature and cross-architecture ability of AdaPruner. 
In contrast, GraNd shows a preference for models on which the gradient norms are computed.
The models used to implement pruning results also have an influence on Glister and Forgetting~\cite{deepcore}.

\subsubsection{Tiny-ImageNet Results}
To further assess the effectiveness of AdaPruner, we prune the Tiny-ImageNet dataset with compression ratios from 70\% to 95\%.
We first resize the Tiny-ImageNet data into $64\times64$, initialize the ResNet-50 model with ImageNet~\cite{imagenet} pre-trained weights, and then fine-tune models using the pruned datasets.
\begin{table}[]
	\centering
        \caption{Test accuracy (\%) of RseNet-50 trained on the pruned dataset of Tiny-ImageNet, along with the corresponding compression ratios.}
	\renewcommand\arraystretch{1.1}
	\resizebox{.95\columnwidth}{!}{
		\begin{tabular}{c|ccccccc }
			\toprule[1.5pt]
			{Method} &70\%&75\%&80\%&85\%&90\%&95\%&100\%\\ \hline
			Random&60.06&68.00&70.66&60.81&64.60&64.67&71.18\\
			GraNd&66.37&68.62&68.45&65.13&72.97&71.50&71.18\\
			EL2N&70.01&64.67&68.71&68.92&71.99&72.67&71.18\\
			Herding&64.00&68.71&70.66&65.79&72.40&71.50&71.18\\
			Glister&70.04&66.05&70.18&67.81&72.90&71.68&71.18\\
			Forgetting&68.26&65.76&68.33&63.20&68.50&73.36&71.18\\
		    CG-Score&66.61&63.31&73.30&65.79&72.24&71.13&71.18 \\
    BNSL&\textbf{70.79}&66.71&72.81&66.83&71.93&71.86&71.18 \\
      Ours&70.02&\textbf{69.28}&\textbf{73.79}&\textbf{70.49}&\textbf{73.00}&\textbf{73.70} &71.18\\
			\bottomrule[1.5pt]
		\end{tabular}
	}
	\label{tiny}
\end{table}

As is shown in Table~\ref{tiny}, AdaPruner consistently achieves higher test accuracy than other dataset pruning methods.
When the datasets are pruned with smaller pruning ratios, such as less than 10\% on Tiny-ImageNet, the removal of redundancies improves the dataset quality and leads to higher accuracy.
Conversely, when higher pruning ratios are applied, the dataset distribution may change, potentially adversely affecting the model performance.
Even under these cases, AdaPruner still introduces the least losses in accuracy, demonstrating its significant effectiveness.
More experimental results on Tiny-ImageNet using Wide-ResNet-50-2~\cite{wrn} can be found in the Appendix.

\subsection{Transfer Learning}
\vspace{-3mm}
\begin{table}[h]
	\centering
        \caption{Transferred test accuracy (\%) on CIFAR-10. }
	\renewcommand\arraystretch{1.}
	\resizebox{.97\columnwidth}{!}{
		\begin{tabular}{c|l|cccccc}
			\toprule[1.5pt]
			\multicolumn{2}{l|}{\textbf{Compression Ratio} (\%)}&60&70&80&90&95&100\\ \hline
			\multirow{8}*{ResNet-18}&Forgetting&87.83&88.55&89.26&89.99&90.53&91.30\\
			&Glister  &88.58 &90.95 &88.39 & 91.43&91.07&91.30 \\
			&GraNd  &71.08 &71.74 &70.32 &69.70&71.56&91.30 \\
			&EL2N  &71.19 & 71.44 & 72.03 &70.46&70.61&91.30 \\
			&Herding &88.58 &88.82 &89.28 &89.86&89.85&91.30 \\
                &CG-Score&90.40&90.88&91.02&91.42&91.10&91.30 \\
                &BNSL&90.07&90.35&90.87&90.97&91.37&91.30 \\                
			&Ours&\textbf{90.50} &\textbf{91.48}& \textbf{91.97}& \textbf{91.92}& \textbf{91.63}&91.30\\ \hline
			\multirow{8}*{ResNet-50}&Forgetting&88.28 & 90.13 &91.04 & 90.66 &91.96&92.34 \\
			&Glister&90.30 &90.42 &91.62 &92.02 &92.19&92.34  \\
			&GraNd & 90.19 &90.89 &90.88 &91.49&92.19&92.34\\
			&EL2N  &90.32 &90.97 &91.61 &91.75&91.70&92.34  \\
			&Herding &90.09 &91.13 & 91.67&91.56&91.77&92.34 \\
                &CG-Score&90.07&91.15&91.52&92.02&\textbf{92.41}&92.34\\
                &BNSL&90.11&90.87&91.80&92.01&91.76&92.34 \\     
			&Ours&\textbf{90.50} &\textbf{91.48}& \textbf{91.97}& \textbf{92.16}& 92.31&92.34\\			
			\bottomrule[1.5pt]
		\end{tabular}
	}
	\label{tab:transferred}
\end{table}
The purpose of dataset pruning aims to improve the training set, thereby enhancing the generalization performance of models.
In addition to the test accuracy, model generalization can also be assessed through transfer learning~\cite{transfer-learning-1,transfer-learning-2}.
To demonstrate transferability, we first pre-train ResNet-18 and ResNet-50 using the pruned CIFAR-100 datasets, followed by fine-tuning the pre-trained models on CIFAR-10.
Table~\ref{tab:transferred} demonstrates that both ResNet-18 and ResNet-50, trained with our pruned datasets, outperform other methods regarding transferred accuracy across various compression ratios. 
These results demonstrate that our approach yields better training datasets and enhances model performance.
\subsection{Object Detection}
\begin{table}[h]
	\centering
        \caption{Detection mAP (\%) results on PASCAL VOC2007 test set. }
	\renewcommand\arraystretch{1.2}
	\resizebox{.95\columnwidth}{!}{
		\begin{tabular}{c|cccccc}
			\toprule[1.5pt]
			\textbf{Compression Ratio} (\%)&75&80&85&90&95&100\\ \hline
			YOLOv3-Tiny&52.97&54.38&55.01&55.24&\textbf{56.65}&56.57\\
			Faster-RCNN&69.26&70.48&70.20&\textbf{70.72}&70.52&70.33 \\
			\bottomrule[1.5pt]
		\end{tabular}
	}
	\label{tab:detection}
\end{table}
This section verifies the scalability of our method to object detection datasets and corresponding models. We conduct experiments on YOLOv3-Tiny~\cite{yolo-tiny} and Faster-RCNN~\cite{faster-rcnn}, both of which use ResNet-50 as the backbone network.
We employ the codebase provided by $Detectron2$~\cite{detectron2} and rigorously follow nearly all the hyperparameters and training strategies.
We first train AdaPruner with YOLOv3-Tiny to obtain the pruned datasets on the union of VOC2007 and VOC2012 \textit{trainval} datasets.
We then employ the pruned datasets to train randomly initialized detector models.
As shown in Table~\ref{tab:detection}, we prune 5\%-25\% data from the training set of object detection models. 
Notably, on removing 5\%-10\% of the training data, models trained with the pruned datasets have achieved consistent improvements over the baseline models trained with the original full dataset.
In particular, we improve the mAP from 56.57\% to 56.65\% on YOLOv3-Tiny, and 70.33\% to 70.72\% on Faster-RCNN, respectively.
Significantly, these enhancements are accompanied by substantial reductions in training overheads.
While most prior works may not be applicable to object detection datasets, we have demonstrated the robust scalability and effectiveness of AdaPruner on object detection datasets and models.

\subsection{Analytical Results}
\subsubsection{Visualization of the Pruning Results on MNIST}
\begin{figure}[h]
	\centering
	\begin{subfigure}[]{0.55\linewidth}
		\centering
		\includegraphics[width=4cm]{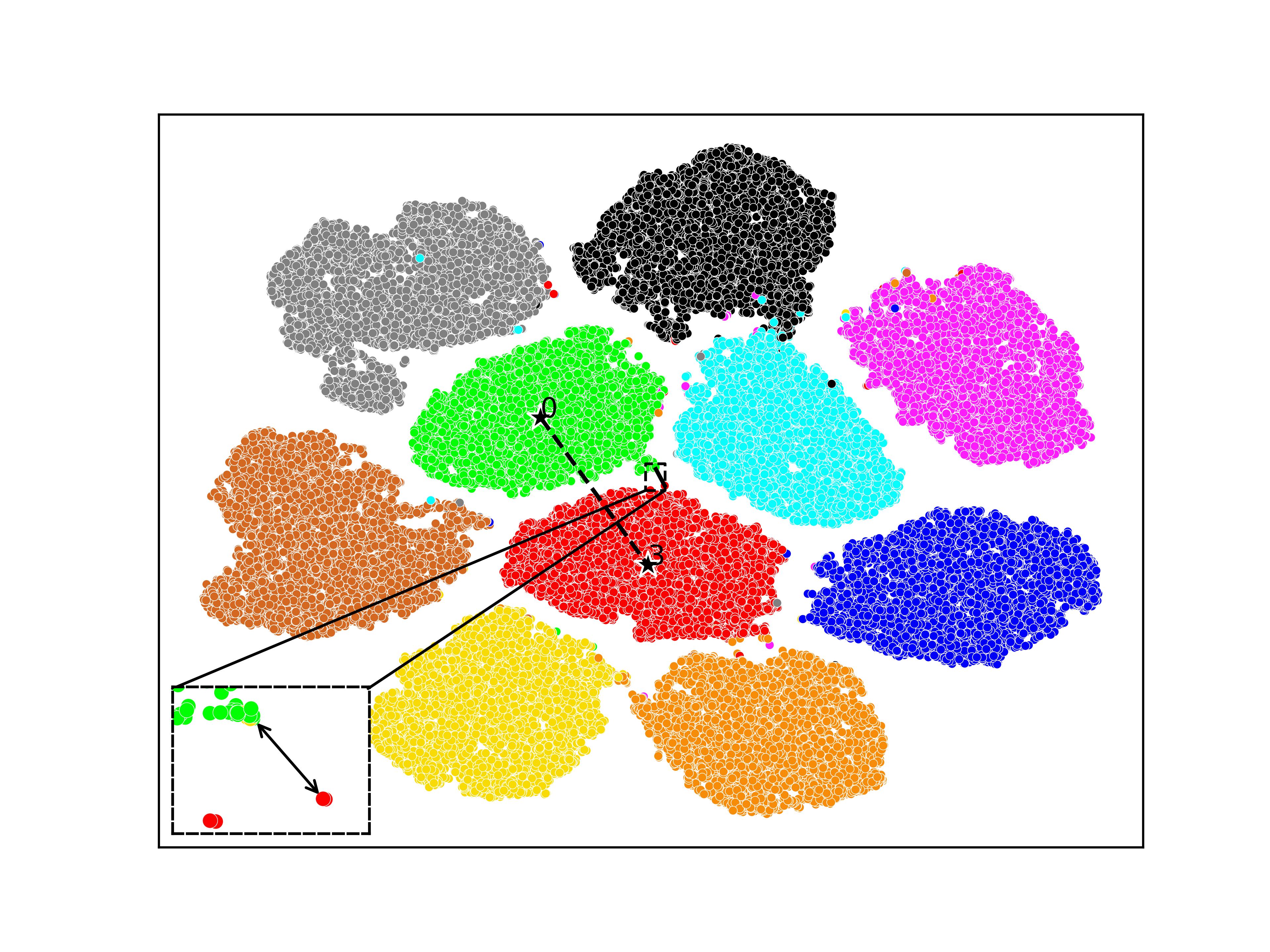}
		\caption{Full dataset using model $M_{full}$.}
		\label{fig2-1}
	\end{subfigure}\\
	\begin{subfigure}[]{0.49\linewidth}
		\centering
		\includegraphics[width=4cm]{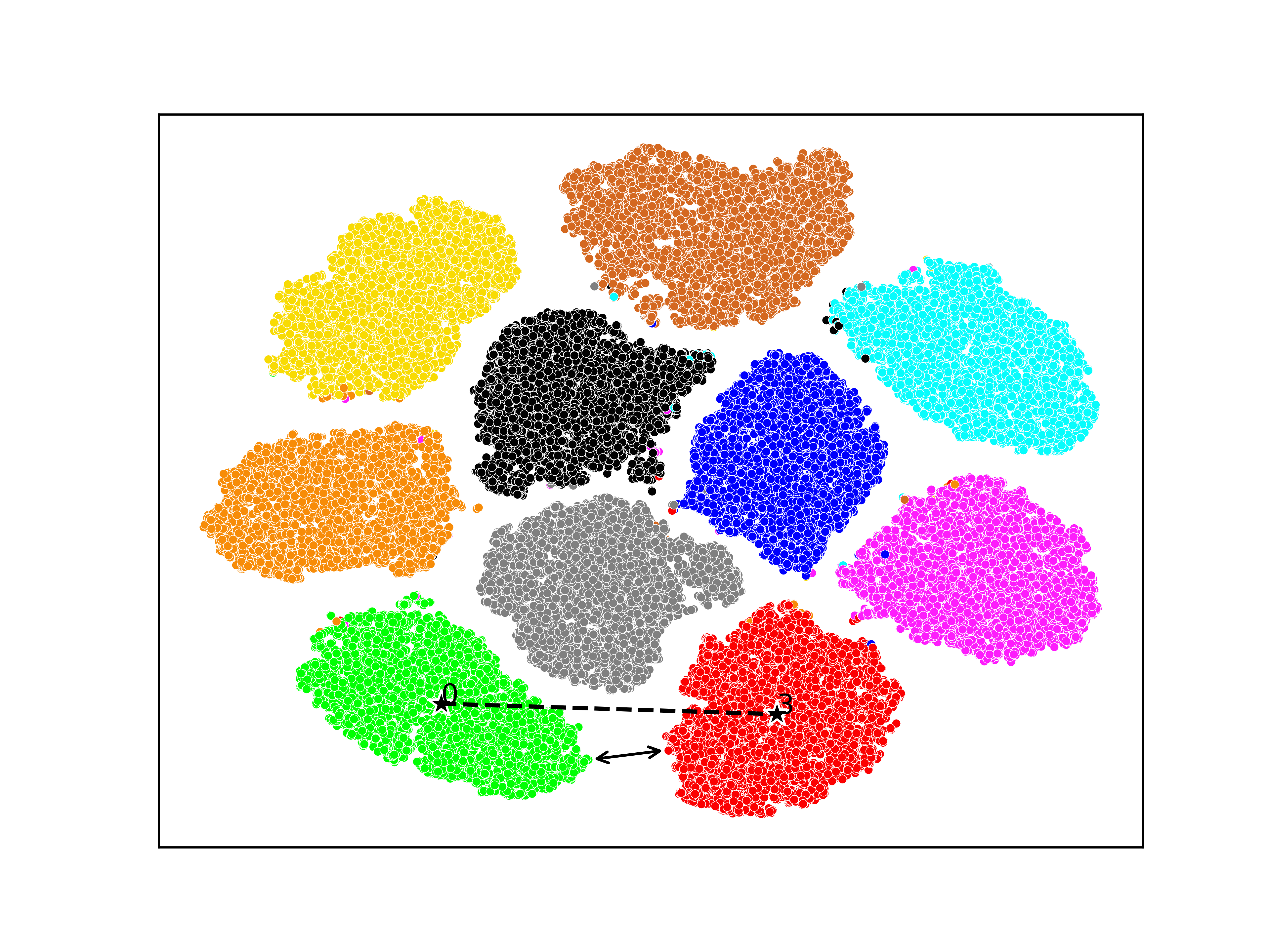}
		\caption{Pruned dataset using model $M_{full}$.}
		\label{fig2-2}
	\end{subfigure}
	\begin{subfigure}[]{0.49\linewidth}
		\centering
		\includegraphics[width=4cm]{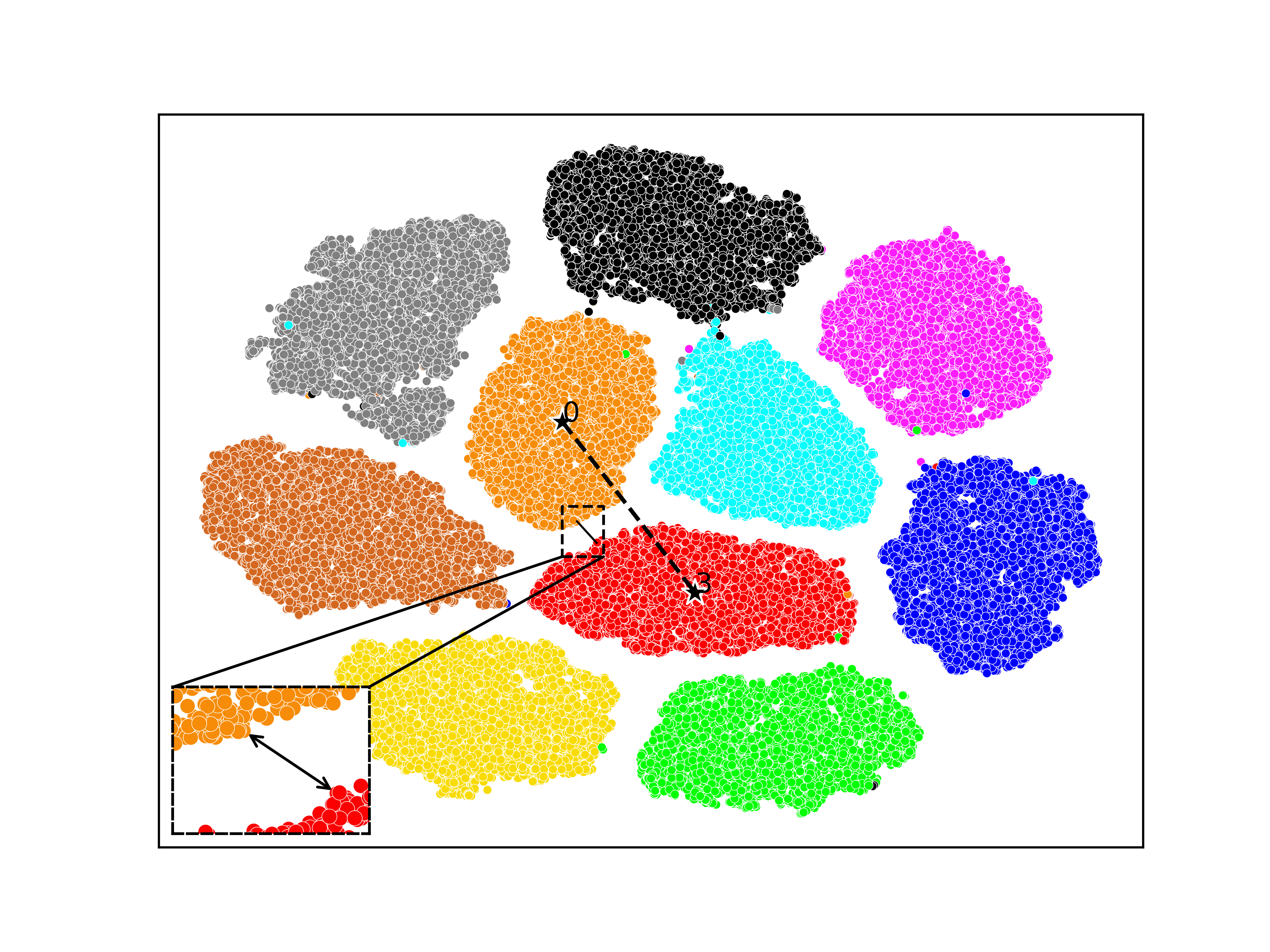}
		\caption{Full dataset using model $M_{pruned}$.}
		\label{fig2-3}
	\end{subfigure}  
	\caption{t-SNE Visualization of the MNIST training set. The ResNet-18 models trained on the full dataset and the pruned dataset (with a 10\% pruning ratio) are denoted as $M_{full}$ and $M_{pruned}$, respectively.}
	\label{fig-tsne}
\end{figure}

To better understand the pruning effects facilitated by AdaPruner, we employ the t-SNE algorithm to visualize the embedding of the MNIST training set in Figure~\ref{fig-tsne}.
We first train a ResNet-18 model on the whole training set and employ it to embed the entire training set, as shown in Figure~\ref{fig2-1}.
Secondly, using our method, we select 90\% of the data from the training set. In Figure~\ref{fig2-2}, we use the same embedding model as Figure~\ref{fig2-1} to illustrate the embedding results of the selected data.
The removal of 10\% of the data significantly transforms the cluster distribution, highlighting t-SNE's sensitivity to input data. 
The pruned embedding results directly obtained from the outcomes of t-SNE are available for comparison in the Appendix.

As depicted in Figure~\ref{fig2-2}, the selected data exhibits an optimized geometric structure.
Specifically, the average distance between each pair of adjacent cluster centers rises by 4.4\%.
The average single linkage (the shortest distance) between each pair of adjacent clusters increases by 17.8\%, while the radius of each cluster (the maximum distance between the data and the corresponding cluster center) decreases by 10.9\%.
To statistically evaluate the clustering results, we utilize the Dunn index (DI)~\cite{dunn2}, which is defined as follows:
\begin{equation}
	D I=\frac{\min _{1 \leq i \neq j \leq m} \delta\left(C_i, C_j\right)}{\max _{1 \leq j \leq m} \Delta_j}
\end{equation}
where the separation metric $\delta\left(C_i, C_j\right)$ is the inter-cluster distance metric between clusters $C_i$ and $C_j$, and the compactness metric $\Delta_j$ calculates the mean distance between all pairs in each cluster.
Therefore, a higher DI means better clustering.
The DIs of the clustering results in Figure~\ref{fig2-1} and Figure~\ref{fig2-2} are $8.79\times10^{-5}$ and $2.05\times 10^{-3}$, respectively.
After removing 10\% of the data, the DI of the clustering results increases significantly.
From an intuitive and statistical standpoint, the clustering results in Figure~\ref{fig2-2} demonstrate \textbf{superior inter-cluster separation and intra-cluster compactness} compared to those in Figure~\ref{fig2-1}.

In Figure~\ref{fig2-3}, the model is trained using only 90\% of the data selected by AdaPruner.
Subsequently, this model is employed to embed the complete training set. 
Although the amount of data in  Figure~\ref{fig2-1} and Figure~\ref{fig2-3} is identical, the embedding models differ. 
Consequently, while the overall distribution in Figure~\ref{fig2-3} exhibits similarity to Figure~\ref{fig2-1}, there are differences in the distribution of individual clusters.
Specifically, the intra-distance decreases by 3.2\%, and the average single linkage between each pair of adjacent clusters increases by 31.16\%.
The DI in Figure~\ref{fig2-3} is $2.59\times10^{-4}$, significantly higher than that in Figure~\ref{fig2-1}.
Thus, the comparison between Figure~\ref{fig2-1} and Figure~\ref{fig2-3} demonstrates that \textbf{models trained using the pruned datasets possess enhanced feature extraction capabilities}.
Therefore, AdaPruner effectively improves datasets and contributes to enhanced model performance.
\subsubsection{Ablation Study}\label{ablation}
\paragraph{Effect of Warm-up Mechanism}
We evaluate the effectiveness of the warm-up mechanism by comparing it with the results obtained by pruning data from scratch.
The mask obtained without the warm-up mechanism is denoted as $Mask_{ww}$.
As shown in Table~\ref{table:ablation-warmup}, compared with the mask used in our method, $Mask_{ww}$ obtains much lower test accuracy on CIFAR-10 under various pruning ratios.
The accuracy gap between ours and $Mask_{ww}$ is as high as 1.46\%.
\begin{table}[]
	\centering
        \caption{Effect of warm-up mechanism on CIFAR-10.    }
	\renewcommand\arraystretch{1.1}
	\resizebox{.47\textwidth}{!}{
		\begin{tabular}{l|c|cccccc}
			\toprule[1.5pt]
			\textbf{\textbf{Model}} & \textbf{Mask} \textbf{Index}& 60\% & 70\%&80\%&90\%&95\%&100\% \\ \hline
			\multirow{2}*{ResNet-18}&Ours&\textbf{95.85}&\textbf{96.29}&\textbf{96.33}&\textbf{96.40}&\textbf{96.42}&96.27\\
			&$Mask_{ww}$& 94.40&95.21 &95.74&95.69 &95.79&96.27\\
            \hline
			\multirow{2}*{ResNet-50}&Ours&\textbf{95.57}&\textbf{96.00}&\textbf{96.25}&\textbf{96.37}&\textbf{96.25}&96.12\\
			&$Mask_{ww}$& 93.65&94.80&95.31  &95.92&95.75&96.12\\  
			\bottomrule[1.5pt]
		\end{tabular}
	}
	\label{table:ablation-warmup}
\end{table}
\vspace{-2mm}
\paragraph{Effect of the Loss Function}
\begin{table}[]
	\centering
        \caption{ Effectiveness of loss terms in pruning CIFAR-10 dataset with 90\% compression ratio.}
	\renewcommand\arraystretch{.9}
	\resizebox{.7\columnwidth}{!}{
		\begin{tabular}{l|cc}
			\toprule[1.5pt]
			 &ResNet-18&ResNet-50 \\ \hline
			w/o $L_{classification}$&96.04 &95.91 \\
			w/o $L_{selection}$&95.87 &96.02  \\
			Full &\textbf{96.37}&\textbf{96.40} \\ 
			\bottomrule[1.5pt]
		\end{tabular}
	}
	\label{tab:ablation-loss}
\end{table}
In Table~\ref{tab:ablation-loss}, we quantitatively assess the impact of each loss term in Eq.~\eqref{loss-function}, where the full loss function yields the best performance across all models.
Without $L_{classification}$, the generalization performance of models deteriorates, and the sample selection degrades into random selection.
Without $L_{comprssion}$, no samples will be selected since models can achieve the lowest selection loss without selecting any samples.
Consequently, we cannot obtain the expected pruning results, and thus we do not report the accuracy without $L_{comprssion}$ in Table~\ref{tab:ablation-loss}.
Without $L_{selection}$, models will incline to prune samples with the highest losses rather than redundant ones.
Consequently, none of the loss terms can be eliminated without compromising the overall performance of the pruning process.
\section{Limitations and Future Work}
While our framework exhibits significant performance across datasets and models, the mask vector used to indicate the selection of each sample might circumscribe its applicability to extremely large-scale datasets, which is also a main challenge for all score-based data pruning approaches.
Future work should therefore emphasize extending existing dataset pruning approaches to such large-scale datasets.
\section{Conclusion}
In this paper, we propose a novel and adaptive end-to-end dataset pruning framework to improve the training datasets and enhance the associated model's performance and efficiency.
Our approach leverages the ADP and PPC modules to iteratively optimize sample selection, thereby achieving effective pruning without the necessity for explicitly defined sample metrics. 
The inherent scalability of the method is emphasized, as it can be applied to any dataset and loss-based deep networks.
Experimental results across various datasets and deep architectures substantiate the superiority of our approach relative to the state-of-the-art, manifesting in improved training datasets as well as enhanced model performance and efficiency.
\bibliography{aaai24}

\end{document}